\documentclass{article}

\usepackage{arxiv}

\usepackage[utf8]{inputenc} 
\usepackage[T1]{fontenc}    
\usepackage{hyperref}       
\usepackage{url}            
\usepackage{booktabs}       
\usepackage{amsfonts}       
\usepackage{amsmath, amssymb}
\usepackage{nicefrac}       
\usepackage{microtype}      
\usepackage{graphicx}
\usepackage{natbib}
\usepackage{doi}
\usepackage{algorithm,algpseudocode}
\usepackage{multirow} 

\title{A Denoising Diffusion-Based Evolutionary Algorithm Framework: Application to the Maximum Independent Set Problem}

\renewcommand{\shorttitle}{A Denoising Diffusion-Based Evolutionary Algorithm}

\hypersetup{
pdftitle={A Denoising Diffusion-Based Evolutionary Algorithm Framework: Application to the Maximum Independent Set Problem},
pdfsubject={cs.LG, cs.NE, cs.AI},
pdfauthor={Joan Salvà Soler, Günther R. Raidl},
pdfkeywords={Denoising Diffusion Models, Evolutionary Algorithms, Combinatorial Optimization, Maximum Independent Set Problem},
}

\author{
  Joan Salvà Soler \\
  Algorithms and Complexity Group\\
  TU Wien\\
  Vienna, Austria \\
  \texttt{joan.salvasoler@h2o.ai} \\
  \And
  Günther R. Raidl \\
  Algorithms and Complexity Group\\
  TU Wien\\
  Vienna, Austria \\
  \texttt{raidl@ac.tuwien.ac.at} \\
}

\begin{document}
\maketitle

\begin{abstract}
    Denoising diffusion models (DDMs) offer a promising generative approach for combinatorial optimization, yet they often lack the robust exploration capabilities of traditional metaheuristics like evolutionary algorithms (EAs).
    We propose a Denoising Diffusion-based Evolutionary Algorithm (DDEA) framework that synergistically integrates these paradigms. It utilizes pre-trained DDMs for both high-quality and diverse population initialization and a novel diffusion-based recombination operator, trained via imitation learning against an optimal demonstrator.
    Evaluating DDEA on the Maximum Independent Set problem on Erdős-Rényi graphs, we demonstrate notable improvements over DIFUSCO, a leading DDM solver. DDEA consistently outperforms it given the same time budget, and surpasses Gurobi on larger graphs under the same time limit, with DDEA's solution sizes being 3.9\% and 7.5\% larger on the ER-300-400 and ER-700-800 datasets, respectively. In out-of-distribution experiments, DDEA provides solutions of 11.6\% higher quality than DIFUSCO under the same time limit. Ablation studies confirm that both diffusion initialization and recombination are crucial.
    Our work highlights the potential of hybridizing DDMs and EAs, offering a promising direction for the development of powerful machine learning solvers for complex combinatorial optimization problems.
\end{abstract}

\keywords{Denoising Diffusion Models \and Evolutionary Algorithms \and Combinatorial Optimization \and Maximum Independent Set Problem}

\setlength{\intextsep}{10pt}     
\setlength{\textfloatsep}{10pt}  

\section{Introduction}
\label{sec:introduction}
Combinatorial Optimization (CO) problems involve finding an optimal solution from a discrete but possibly huge set of candidates. Many CO problems are NP-hard and challenging to solve in practice for practically relevant instances. Traditional approaches to tackle such problems often rely on methods like Integer Linear Programming (ILP) or carefully designed (meta-)heuristics, which often require substantial domain expertise and may struggle with scalability.

Advances in deep learning have led to new methods for tackling CO problems~\cite{bengio2021mlopt}, with Graph Neural Networks (GNNs) emerging as a foundational architecture for many of these approaches, owing to the inherently graph-structured nature of many CO problems \cite{jin2024uni}. More recently, generative models like Denoising Diffusion Models (DDMs) have demonstrated competitive performance on CO problems, reaching state-of-the-art performance among neural solvers for the Traveling Salesman Problem (TSP) \cite{difusco,yu2024disco,li2023t2t}. Frameworks such as DIFUSCO \cite{difusco} aim to efficiently generate diverse, high-quality candidate solutions in parallel by leveraging DDMs trained in a supervised manner from solved instances. However, these diffusion-based methods still face significant challenges, including ensuring the feasibility of generated solutions and generalizing effectively to different problem sizes or variants. These methods are often reliant on problem-specific post-processing heuristics to achieve competitive performance.

This reliance on subsequent, often problem-specific, refinement stages highlights a critical gap. While DDMs excel at efficiently generating potentially diverse solutions, they frequently lack the robust, structured exploration mechanisms inherent in classical metaheuristics. Conversely, metaheuristics like evolutionary algorithms (EAs) provide a general framework for robust search but can struggle without effective methods to initialize populations with high-quality, diverse individuals or to generate promising offspring through recombination and mutation. Integrating the generative strengths of DDMs with the structured search of EAs presents an opportunity to overcome the individual limitations.

To bridge this gap, we propose the Denoising Diffusion-Based Evolutionary Algorithm (DDEA), a novel hybrid framework that integrates pre-trained DDMs into core operators of an EA. Specifically, we (1) use the DDM to initialize a diverse population of high-quality solutions for exploration and (2) train a novel diffusion-based recombination operator using imitation learning from an ILP-based demonstrator for exploitation. This synergy aims to combine the efficient generation of diverse, high-quality solutions from DDMs with the problem-agnostic exploratory capabilities of EAs. Our DDEA goes beyond using DDMs merely as a solution generator followed by refinement. Instead, the refinement is embedded within the evolutionary loop, using DDM inference to improve the quality of the offspring. We demonstrate and experimentally evaluate the effectiveness of the DDEA on the Maximum Independent Set Problem (MISP).

The remainder of this paper is structured as follows: Section~\ref{sec:related-work} discusses related work. Section~\ref{sec:ddea} details the proposed DDEA framework. Section~\ref{sec:experiments} presents the experimental setup and results. Finally, Section~\ref{sec:conclusion} concludes the paper and discusses future work. 

\section{Related Work}
\label{sec:related-work}

This section reviews literature relevant to machine learning for combinatorial optimization, with a focus on denoising diffusion-based methods and DIFUSCO. We also briefly review evolutionary algorithms and hybrid ML-EA approaches.

Machine Learning approaches for CO can be broadly categorized along two axes: the type of learning employed and the nature of algorithmic integration with CO frameworks \cite{bengio2021mlopt}.

From the learning type perspective, imitation learning trains models to mimic expert functions, often used to approximate expensive components like MIP branching decisions \cite{khalil2016learning}. Reinforcement learning (RL) enables agents to learn policies through trial-and-error interaction with the optimization environment, successfully applied to problems like the traveling salesman problem (TSP) \cite{kool2018attention}. Gradient-based learning uses neural networks to parameterize the solution space and optimizes directly using continuous relaxations, representing an unsupervised approach \cite{karalias2020erdos,sanokowski2024diffusion}.

From the algorithmic integration perspective, end-to-end learning trains a single ML model to map problem instances directly to solutions, exemplified by pointer networks for TSP \cite{vinyals2015pointer}, autoregressive approaches like transformer-based models~\cite{kool2018attention} and generative flow network models~\cite{zhang-23a}, and denoising diffusion-based models like DIFUSCO \cite{difusco}. Another paradigm is learning to configure algorithms, which uses ML to tune parameters or select heuristics for existing CO solvers, such as predicting good MIP solver settings \cite{kemminer2024configuring} or using LLMs to propose heuristic components \cite{ye2024reevo}. Finally, ML alongside optimization algorithms involves tight, iterative coupling where the ML model guides the CO solver's decisions, like predicting branching variables or node selection strategies in branch-and-bound \cite{scavuzzo2024machine}.

Across many of these ML approaches, particularly those tackling problems with inherent graph structures, graph neural networks (GNNs) have become a foundational architectural component \cite{jin2024uni,georgiev2024neural}. Their capacity to process graph data efficiently allows them to learn representations that effectively capture the combinatorial relationships within CO problem instances, making them suitable for diverse tasks from RL policy learning \cite{khalil2017learning} to end-to-end solution generation \cite{difusco,karalias2020erdos}.
Many CO problems have an inherent graph structure, which makes GNNs a natural fit for these problems. The work by \cite{georgiev2024neural} surveys the use of GNNs in CO, while \cite{jin2024uni} proposes a unified GNN framework for CO problems that includes techniques for graph representation, converting non-graph problems to graph form, graph decomposition, and graph simplification to enable efficient GNN-based solutions.

\paragraph{Denoising Diffusion Models for Combinatorial Optimization.}

Denoising Diffusion Models (DDMs) \cite{ho2020denoising} represent a powerful class of generative models that are primarily known from applications in image and video generation. They are trained by adding noise to labeled training data (forward process) and learning to reverse this process (backward process, i.e., denoising) in a stepwise fashion. A trained model is then iteratively applied to random noise to generate new samples following the probability distribution of the training data.
Sun and Yang \cite{difusco} pioneered the use of DDMs for CO with DIFUSCO, proposing a denoising diffusion model powered by a GNN that can generate high-quality solutions for problems like the TSP and the MISP. Key characteristics include its potential to capture multi-modal solution distributions and faster sampling compared to autoregressive models. However, DIFUSCO often requires problem-specific post-processing heuristics (e.g., greedy decoding, local search) to ensure feasibility and enhance solution quality.
In comparison to autoregressive generative models, denoising diffusion models like DIFUSCO have the advantage of potentially scaling much better to larger instances due to the parallel nature of their solution construction when making use of an efficient GNN implementation on a GPU. Moreover, many diverse solution candidates can be generated in parallel when a powerful-enough GPU is provided.

Subsequent works have sought to improve DIFUSCO. With DISCO, Yu et al.~\cite{yu2024disco} accelerate inference by using residue-guided sampling and analytical denoising steps. The work of \cite{huang2023accelerating} employs progressive distillation to compress the diffusion process into fewer steps, achieving substantial speedups while maintaining high solution quality. Others explore alternative learning paradigms; Li et al.~\cite{li2023t2t} combine supervised diffusion learning with an unsupervised gradient-based search during inference, which can improve the quality of the generated solutions.
The work of \cite{sanokowski2024diffusion} introduces a purely unsupervised diffusion framework that allows training without labeled optimal solutions.

\paragraph{Evolutionary Algorithms.}

Evolutionary algorithms \cite{back1997handbook} are metaheuristics inspired by natural evolution. They maintain a population of candidate solutions, iteratively improving them through selection, recombination (crossover), and mutation. 
Their application to the MISP has been explored using different approaches, such as permutation-based encodings combined with greedy decoders and operators like the partially matched crossover \cite{back1996evolutionary}, or direct binary representations using penalty functions to handle constraints and standard genetic operators like two-point crossover and bit-flip mutation \cite{liu1996genetic}.

\paragraph{ML-EA Hybridization.}

The integration of machine learning (ML) with EAs has emerged as a powerful paradigm for solving complex optimization problems. Recent surveys, such as \cite{Liang2024}, systematically analyze surrogate-assisted evolutionary algorithms that employ ML models to approximate expensive fitness evaluations, significantly reducing computational costs. Beyond surrogate assistance, ML techniques have been integrated into EA operators, including adaptive parameter control strategies and reinforcement learning-guided search mechanisms \cite{bai2023evolutionary}.
A major departure from traditional EA paradigms is the Learnable Evolution Model (LEM) \cite{Michalski2000}, which replaces stochastic operators with inductive learning to generate targeted offspring, enabling ``quantum leaps'' in fitness.
\section{The DDEA Framework}
\label{sec:ddea}

This section details DDEA, our proposed framework integrating pre-trained DDMs into the classical EA cycle. The primary motivation is to leverage the powerful generative capabilities of DDMs, which learn rich distributions over complex solution spaces, to enhance the exploration and exploitation abilities of EAs for challenging combinatorial optimization problems. 

The core idea of DDEA is not to replace the entire EA but to augment or substitute specific components—namely initialization and recombination—with operations guided by a DDM pre-trained on the target problem distribution. This hybrid approach aims to combine the global search perspective inherent in EAs with the high-quality, structure-aware sampling provided by DDMs. 

\begin{figure}[t!]
    \centering
    \includegraphics[width=\textwidth, trim=0.1cm 0.15cm 0cm 0.5cm, clip]{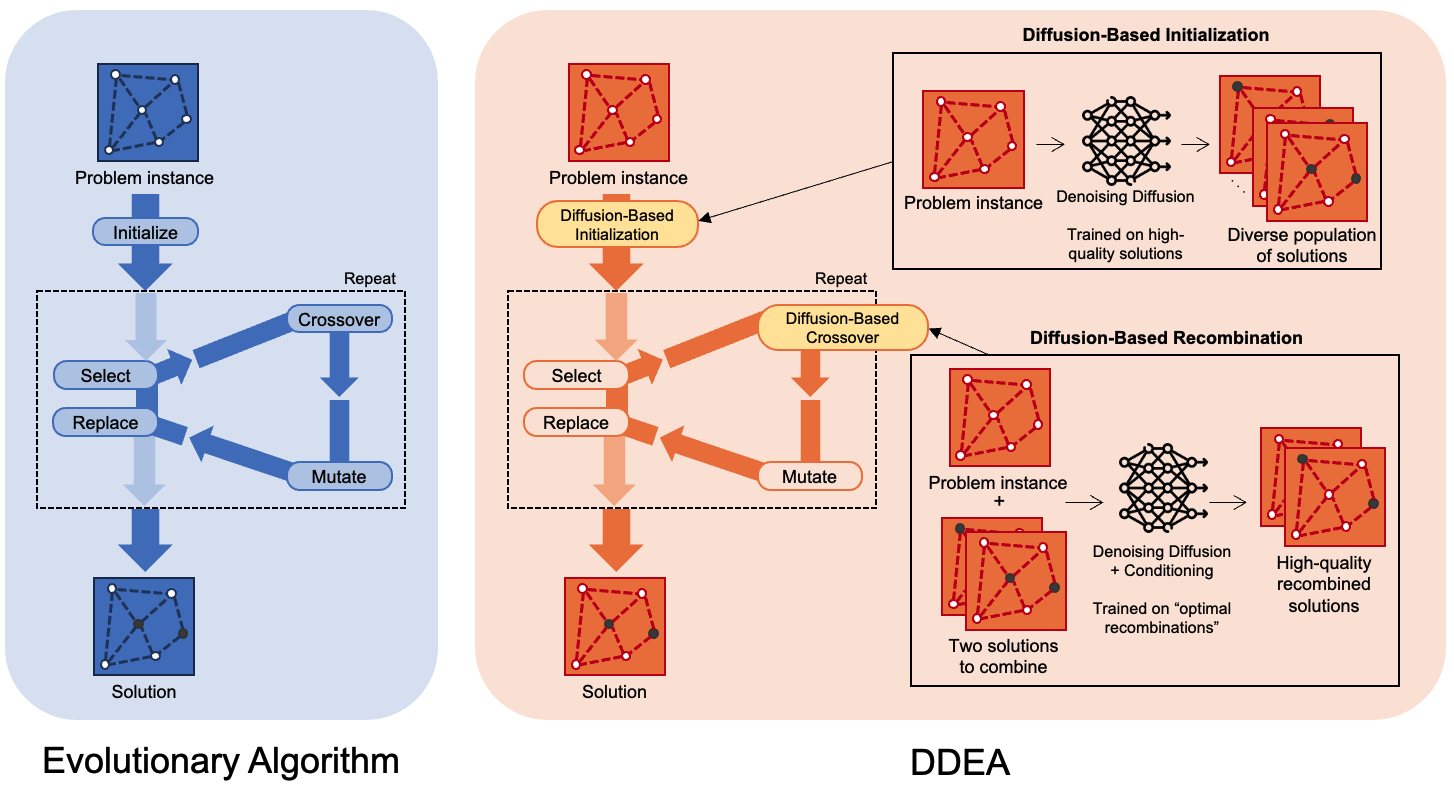}
    \caption{Overview of the DDEA design in contrast to a classical EA. DDEA replaces standard initialization and recombination with diffusion-based counterparts.}
    \label{fig:dea_overview}
\end{figure}

The DDEA retains a standard EA loop. Parent solutions are selected by binary tournaments, the diffusion-based recombination operator is applied, followed by a standard mutation operator, and we replace with elitism, never accepting duplicate solutions.
Figure \ref{fig:dea_overview} provides a high-level comparison between the workflow of a standard EA and the DDEA.

The key strength of DDEA lies in its problem independence -- the framework can be applied to any combinatorial optimization problem with minimal adaptations. Only two problem-specific components are required: a simple decoding strategy to transform probability vectors into feasible solutions, and a source of training labels (which can be obtained from any specialized solver).

\subsection{The Maximum Independent Set Problem, its Solution Representation, and Solution Decoding}
\label{sec:mis_decode}

We use the MISP here as an exemplary problem to illustrate the DDEA framework and to make concrete experiments.
This problem is defined on an undirected graph $G=(V,E)$, and its goal is to find the largest subset of vertices $S \subseteq V$ such that no two vertices in $S$ are adjacent. Formally, we want to maximize the size of $S$ while ensuring that for all pairs of vertices $u, v \in S$, $(u,v) \notin E$.
We encode candidate solutions as a binary incidence vector of size $n=|V|$, where each element $x_i$ represents the presence (1) or absence (0) of node $i\in V$ in the independent set. 

A problem-specific component of our framework is the decoding process that transforms a vector of node probabilities obtained from a DDM into a feasible solution. This decoding process operates as follows. Given an input probability vector $\mathbf{p} \in [0,1]^n$, which represents the likelihood of each node to be in the independent set, we first sort all nodes in descending order of their probability scores. Starting with an empty solution, we then process nodes in this order. For each node $i$, if it has not been marked as ineligible, we include it in our solution and mark all its neighbors as ineligible (since including any neighbor would violate the independence constraint). This greedy construction ensures that high-probability nodes are prioritized while maintaining feasibility at each step. The resulting solution is always a valid independent set, making the population remain feasible at all times. This decoding algorithm can be vectorized over the population, achieving sublinear time growth in the population size $P$.

The DDEA employs a standard mutation operator, which is applied with a fixed probability $p_m$ to each solution in the population. If a solution $x$ is selected for mutation, each vertex selected in the current solution ($x_i=1$) is deselected with probability $p_d$. The mutated solution $x'$ is the result of decoding a probability vector $\mathbf{p}$. This vector $\mathbf{p}$ is sampled from $\mathcal{U}(0,1)^n$, with the modification that $\mathbf{p}[i]$ is forced to $0$ if $x_i = 1$ and vertex $i$ is deselected, and forced to $1$ if $x_i = 1$ and vertex $i$ is not deselected.

\subsection{Diffusion-Based Initialization}

To generate a high-quality and diverse initial population, DDEA leverages a pre-trained DIFUSCO model \cite{difusco}. Crucially, DIFUSCO exhibits the multi-modal property \cite{difusco,yu2024disco}, enabling it to sample diverse solutions from the underlying distribution when starting from different noise. This makes it well-suited for initializing an EA population.

The initialization involves two steps. First, we sample node probability vectors (heatmaps) $\mathbf{p}_j \in [0,1]^n$, $j=1,\ldots,P$ from the pre-trained DIFUSCO model for the target MISP efficiently in parallel, where $P$ denotes the population size. Second, we decode all these heatmaps $\mathbf{p}_j$ into feasible candidate solutions $x_j \in \{0, 1\}^n$ using the above described decoding heuristic, ensuring all initial individuals are valid independent sets.

\subsection{Diffusion-Based Recombination}
\label{subsec:diffusion_recombination}

Recombination (or crossover) in genetic algorithms is the major operator to generate promising offspring by inheriting and combining properties of usually two parental candidate solutions. Traditional crossover operators (like uniform or single-point crossover) are often generic and may not effectively preserve or combine beneficial structural information specific to the problem domain (e.g., graph structures in MISP).

The goal of diffusion-based recombination is to be a more intelligent and problem-aware, learned operator. Instead of simple bit mixing in a naive randomized fashion, we use a conditional diffusion model to generate offspring. This model is trained to sample high-quality solutions from a distribution conditioned on the provided parent solutions, in the following labeled $x$ and $y$. The intuition is that the model learns how to effectively merge high-quality features of both parents within the learned manifold of high-quality MISP solutions.

\paragraph{Optimized Recombination as an Expert Demonstrator.}

To train our diffusion recombination model, we require training samples of high-quality offspring corresponding to given parent pairs. We generate this training data using an \emph{expert demonstrator} -- an operator designed to produce high-quality offspring, even if computationally expensive. This demonstrator, that we refer to as ``optimized recombination'', is realized by means of an Integer Linear Program (ILP). It tries to find the best solution in a restricted search space centered around two parent solutions. Let $x, y \in \{0,1\}^n$ be the parent solutions for a MISP instance given by a graph $G=(V,E)$, and $z \in \{0,1\}^n$ be the decision variables for the offspring. Let $S_v = \{i \in V \mid v_i = 1\}$ be the solution set for a given incidence vector $v$ and $U_v = V \setminus S_v$ be the corresponding unselected nodes. The ILP is defined as follows:

{\small
\begin{align}
    \max \ \ & \sum_{i \in V} z_i \label{eq:mis_objective}\\
    \text{s.t.} \ \ & z_u + z_v \leq 1 & \forall (u, v) \in E \label{eq:mis_constraint1}\\
    & \underbrace{\left( \sum_{i \in S_x} (1 - z_i) + \sum_{i \in U_x} z_i \right)}_{h(z, x)} + \underbrace{\left( \sum_{i \in S_y} (1 - z_i) + \sum_{i \in U_y} z_i \right)}_{h(z, y)} \leq k \label{eq:mis_constraint2}\\
    & k = \lambda \cdot \left( \sum_{i \in S_x} (1 - y_i) + \sum_{i \in U_x} y_i \right) \label{eq:mis_constraint3}\\
    & z_i \in \{0, 1\} & \forall i \in V \label{eq:mis_constraint4}
\end{align}
}

The objective function \eqref{eq:mis_objective} maximizes the offspring independent set size. Constraints \eqref{eq:mis_constraint1} and \eqref{eq:mis_constraint4} enforce offspring feasibility. Inspired by local branching \cite{fischetti2003local}, constraint \eqref{eq:mis_constraint2} limits the combined Hamming distance $h(z, x) + h(z, y)$ between child $z$ and parents $x, y$ to $k$. The threshold $k$ is defined in \eqref{eq:mis_constraint3} as a factor $\lambda$ of the parent distance $h(x,y)$. Parameter $\lambda$ controls exploration: larger values allow more deviation from parents while smaller values restrict the search space. Solving this ILP with an exact solver (subject to a time limit) yields a high-quality offspring $z^*$, which serves as the expert target for training the diffusion model.

\paragraph{The Diffusion Recombination Model Architecture.}

To enable parent conditioning in the recombination model, we extend DIFUSCO's input features to include each node's status in both parent solutions $x$ and $y$ alongside its positional embedding. The three features are independently embedded to dimension $d_h$ and summed to form the initial node representation, following \cite{vaswani2017attention}. This modified input allows the GNN layers to process the graph structure concurrently with parent solution information during denoising steps. The remaining diffusion architecture (GNN layers, time embedding, output layers) and the denoising process follow the practices outlined in \cite{difusco}.

\paragraph{Training via Imitation Learning.}

In order to train using imitation learning, we first generate a dataset of expert demonstrations.
We consider a DDEA variant where we apply the optimized recombination operator, i.e., the ILP from above, as the recombination.
During runs of this DDEA variant, the ILP recombination operator is invoked to compute the expert offspring $z^*$ for each selected parent pair $(x, y)$, and the resulting triplets $(x, y, z^*)$, along with the corresponding graph $G$, are stored to form the training dataset.
This dataset then forms the basis for training the diffusion model.
Its generation is computationally expensive, but we argue that it only needs to be done once offline on representative instances, so this runtime is not a significant concern.
More specifically, a GNN model is trained to iteratively predict for two given parents $x$ and $y$ and some initial uniform noise how to reduce noise to ultimately obtain $z^*$, conditioned on time steps and the graph $G$ besides the parent solutions $x$ and $y$.

During the recombination phase of the DDEA, given a selected parent pair $(x, y)$, we use the trained conditional diffusion model to sample one or more offspring heatmaps $\mathbf{p}$. These heatmaps are then converted into feasible offspring by again applying the decoding heuristic. This approach allows DDEA to generate new solutions that are not random combinations of parents but are guided by the learned distribution of high-quality solutions conditioned on those parents.

\section{Experimental Evaluation}
\label{sec:experiments}

We evaluate our DDEA framework on the MISP using Erdős-Rényi (ER) random graphs of varying sizes.
Experiments focus on evaluating the core diffusion components, performing ablation studies, benchmarking against other methods, and evaluating out-of-distribution generalization.

\subsection{Setup}
\label{sec:exp_setup}

\paragraph{Datasets.} We use Erdős-Rényi (ER) graphs generated as in \cite{boether2022whats} using edge probability $0.15$. Three sets of graphs were generated with nodes $n$ sampled from $[50, 100]$, $[300, 400]$, and $[700, 800]$. For each range, 40000 graphs were created, keeping 128 for testing. For ER-700-800, we reuse the testing dataset of Sun and Yang~\cite{difusco} and only generate the training set.
Ground truth labels are obtained using KaMIS \cite{lamm2017finding} with a 60\,s time limit. We use an Out-of-Distribution dataset of 42 ER graphs with $n$ from $[1300, 1400]$ for generalization tests.

\paragraph{Implementation.} Our DDEA implementation uses PyTorch Geometric and EvoTorch, with DDMs based on DIFUSCO~\cite{difusco}. We used Gurobi\footnote{\url{https://www.gurobi.com}} 12.0 for the ILP-based optimized recombination operator in single-threaded mode with a 15\,s time limit.
DDM inference and training used an NVIDIA A100 80\,GB GPU. Other experiments used a cluster with Intel(R) Xeon(R) E5-2640 v4 CPU at 2.40\,GHz and 160\,GB RAM on Ubuntu 18.04.6 LTS.

\paragraph{DDEA Configuration.} We defer the discussion on the population size $P$ and the number of generations $G$ to the next section.
The mutation probability is $p_m = 0.1$ and the deselection probability $p_d = 0.05$. Moreover, for the allowed deviation from parental solutions in expert recombination training, we set $\lambda = 1.75$. These parameters were identified by preliminary experiments and limited tuning.
The DDMs use DIFUSCO's anisotropic GNN architecture with edge gating, hidden dimension $d_h = 256$ and $n_l=12$ layers. Following \cite{difusco}, we use Gaussian diffusion for ER graphs.
For inference, we use a cosine schedule with $T_\mathrm{inf}=50$ denoising steps. For training, we use a linear schedule with $T_\mathrm{train}=1000$ steps and cosine inference schedule, batch size 64, learning rate 0.0002, and cosine-decay with weight decay 0.0001, as in~\cite{difusco}.

\paragraph{Metrics.} We measure solution quality using the gap to the best-known solution (BKS): $\text{\%-Gap} = 100\%\cdot \frac{\textrm{BKS-size}-\textrm{solution-size}}{\textrm{BKS-size}}$. 
For final benchmarking, we follow \cite{difusco} and report mean cost. We also use the Primal Integral~\cite{berthold2021}, normalized by the area of the rectangle defined by the best-known solution value: $\frac{\sum_{i=1}^{n} a_{i-1} \cdot (t_i - t_{i-1})}{(t_n - t_0) \cdot \textrm{BKS-size}}$, where $a_i$ is the incumbent solution value at time $t_i$. 

\paragraph{Baselines for the recombination and initialization operators.}
\label{par:baselines}
For ablation, we use baseline operators to serve as non-diffusion counterparts. For initialization, we use a randomized greedy heuristic (RG) generating probability vectors decoded via the greedy heuristic from Section~\ref{sec:mis_decode}. For recombination, we use the Consensus-Divergence Crossover (CDX), which takes two parents, identifies common selected vertices, and generates two offspring probability vectors. The first offspring assigns probability one to vertices selected in both parents to strongly encourage their selection (consensus), while the second offspring assigns probability zero to these common vertices to encourage diversity (divergence). We finally decode the offspring using the greedy decoding.

\subsection{Evaluation of the Diffusion-Based Recombination}
\label{sec:eval_recomb}

We first evaluate the diffusion recombination operator in isolation. It is trained on $\approx$160000 examples from expert ILP recombination across all three datasets. Table \ref{tab:recomb_eval} compares the gaps to the expert recombination label against DIFUSCO x2 inference and CDX baseline using parents of varying quality: high-quality EA parents, medium-quality heuristic solutions from decoded random vectors, and low-quality random solutions. The results show that the diffusion recombination effectively approximates the expert operator, outperforming CDX and DIFUSCO. Performance degrades with decreasing parent quality, demonstrating cost monotonicity and validating the design of the operator.

\begin{table}[t!]
    \centering
    \renewcommand{\arraystretch}{0.86}
    \caption{Average gap (\%) to the optimized recombination label for different approaches and parent types.}
    \label{tab:recomb_eval}
    \setlength{\tabcolsep}{3.5pt}
    \begin{tabular}{ll*{3}{r}}
        \toprule
        \multirow{2}{*}{Method} & \multirow{2}{*}{Parent Type} & \multicolumn{3}{c}{Dataset / Gap (\%)} \\
        \cmidrule(lr){3-5}
        & & ER-50-100 & ER-300-400 & ER-700-800 \\
        \midrule
        \multirow{3}{*}{\shortstack[l]{Diffusion\\Recombination}} & EA Parents & \textbf{1.28 $\pm$ 3.34} & \textbf{6.10 $\pm$ 8.53} & \textbf{5.18 $\pm$ 8.48} \\
        & Heuristic & 2.86 $\pm$ 4.40 & 12.18 $\pm$ 6.86 & 16.81 $\pm$ 6.38 \\
        & Random & 3.10 $\pm$ 5.09 & 26.28 $\pm$ 5.90 & 26.60 $\pm$ 4.15 \\
        \midrule
        {DIFUSCO} & -- & 6.36 $\pm$ 5.54 & 8.51 $\pm$ 7.17 & 5.44 $\pm$ 4.89 \\
        \midrule
        {CDX Recombination} & EA Parents & 6.38 $\pm$ 7.25 & 20.52 $\pm$ 7.61 & 24.13 $\pm$ 6.03 \\
        \bottomrule
    \end{tabular}
\end{table}

\subsection{Evaluating DDEA Components within the EA Framework}
\label{sec:ablation}

Table~\ref{tab:ea_recombination_comparison} assesses the contribution of the diffusion-based initialization (DI) and recombination (DR) within the complete EA framework. We compare DDEA ($P=16,\ G=20$) against the following variants:
\vspace{-0.5em}
\begin{itemize}
    \item DDEA without DR: Uses DI with CDX recombination (DI / CDX)
    \item DDEA without DI: Uses DR with RG initialization (RG / Diffusion)
    \item Naïve EA: Uses RG initialization and CDX recombination (RG / CDX)
    \item Expert Recombination EA: Uses DI and ILP recombination (DI / Optim.)
\end{itemize}
\vspace{-0.5em}
The table shows gaps, hyperparameters, total runtime $t$ and runtime per generation $t/\mathrm{gen}$.
The results show that full DDEA achieves performance comparable to expert optimized recombination, with minimal gap differences. The learned diffusion recombination effectively approximates optimized recombination while drastically reducing runtime per generation (e.g., 5\,s vs.~88\,s on ER-300-400). The full DDEA significantly outperforms the Naïve EA variants (RG/CDX and DI/CDX) under matched time limits. Comparing the full DDEA to the DI/CDX variant highlights the substantial benefit of diffusion recombination over classical methods (e.g., 2.81\% vs.~5.75\% gap on ER-700-800). Similarly, comparing the full DDEA to the RG/DR variant shows the importance of DI (e.g., 0.94\% vs.~1.39\% gap on ER-300-400). Although diffusion-based initialization is beneficial, the diffusion recombination operator demonstrates robustness, performing well even with greedy initialization (RG/DR) compared to the Naïve baseline (RG/CDX). While the runtime of DDEA is dominated by the diffusion steps during recombination, the learned operator successfully approximates the expensive ILP solving, enabling near-expert performance in more practical runtimes.

\begin{table}[t!]
    \centering
    \renewcommand{\arraystretch}{0.80}
    \caption{Performance comparison of different EA variants across datasets.}
    \setlength{\tabcolsep}{4pt}
    \begin{tabular}{lllrrrrr}
        \toprule
        Dataset & Init. & Recomb. & {$G$} & {$P$} & \%-Gap & $t$ [s] & $t/\mathrm{gen}$ [s] \\
        \midrule
        \multirow[c]{6}{*}{ER-50-100} & DI & \multirow[c]{2}{*}{CDX} & 50 & 16 & $0.55 \pm 1.52$ & $24.1 \pm 7.5$ & $0.5 \pm 0.2$ \\
         & RG &  & 50 & 50 & $0.25 \pm 1.15$ & $34.6 \pm 5.1$ & $0.7 \pm 0.1$ \\
        \cmidrule{2-8}
         & DI & Optim. & 20 & 16 & $0.00 \pm 0.00$ & $11.4 \pm 4.0$ & $0.6 \pm 0.2$ \\
        \cmidrule{2-8}
         & DI & \multirow[c]{2}{*}{Diffusion} & 20 & 16 & $0.11 \pm 0.53$ & $14.4 \pm 1.8$ & $0.7 \pm 0.1$ \\
         & RG &  & 20 & 16 & $0.11 \pm 0.69$ & $12.5 \pm 1.9$ & $0.6 \pm 0.1$ \\
        \midrule
        \multirow[c]{6}{*}{ER-300-400} & DI & \multirow[c]{2}{*}{CDX} & 75 & 16 & $3.57 \pm 2.27$ & $76.6 \pm 10.6$ & $1.0 \pm 0.1$ \\
         & RG &  & 75 & 16 & $8.58 \pm 2.98$ & $48.5 \pm 4.5$ & $0.6 \pm 0.1$ \\
        \cmidrule{2-8} 
         & DI & Optim. & 5 & 16 & $0.58 \pm 1.25$ & $441.4 \pm 64.6$ & $88.3 \pm 12.9$ \\
        \cmidrule{2-8}
         & DI & \multirow[c]{2}{*}{Diffusion} & 20 & 16 & $0.90 \pm 1.37$ & $188.0 \pm 16.9$ & $9.4 \pm 0.8$ \\
         & RG &  & 20 & 16 & $1.39 \pm 2.09$ & $109.0 \pm 27.3$ & $5.4 \pm 1.4$ \\
        \midrule
        \multirow[c]{6}{*}{ER-700-800} & DI & \multirow[c]{2}{*}{CDX} & 305 & 16 & $5.75 \pm 2.66$ & $231.7 \pm 33.5$ & $0.8 \pm 0.1$ \\
         & RG &  & 305 & 16 & $13.36 \pm 3.48$ & $238.0 \pm 57.0$ & $0.8 \pm 0.2$ \\
        \cmidrule{2-8}
         & DI & Optim. & 3 & 16 & $3.21 \pm 2.09$ & $314.4 \pm 27.3$ & $104.8 \pm 9.1$ \\
        \cmidrule{2-8}
         & DI & \multirow[c]{2}{*}{Diffusion} & 20 & 16 & $2.95 \pm 2.02$ & $478.2 \pm 9.5$ & $23.9 \pm 0.5$ \\
         & RG &  & 20 & 16 & $4.52 \pm 2.33$ & $473.7 \pm 37.1$ & $23.7 \pm 1.9$ \\
        \bottomrule
    \end{tabular}
    \label{tab:ea_recombination_comparison}
\end{table}

\subsection{Comparing to Other Approaches for the MISP}
\label{sec:benchmarking}

\begin{figure}[t!]
    \centering
    \label{fig:cost_vs_budget}
    \includegraphics[width=1\textwidth]{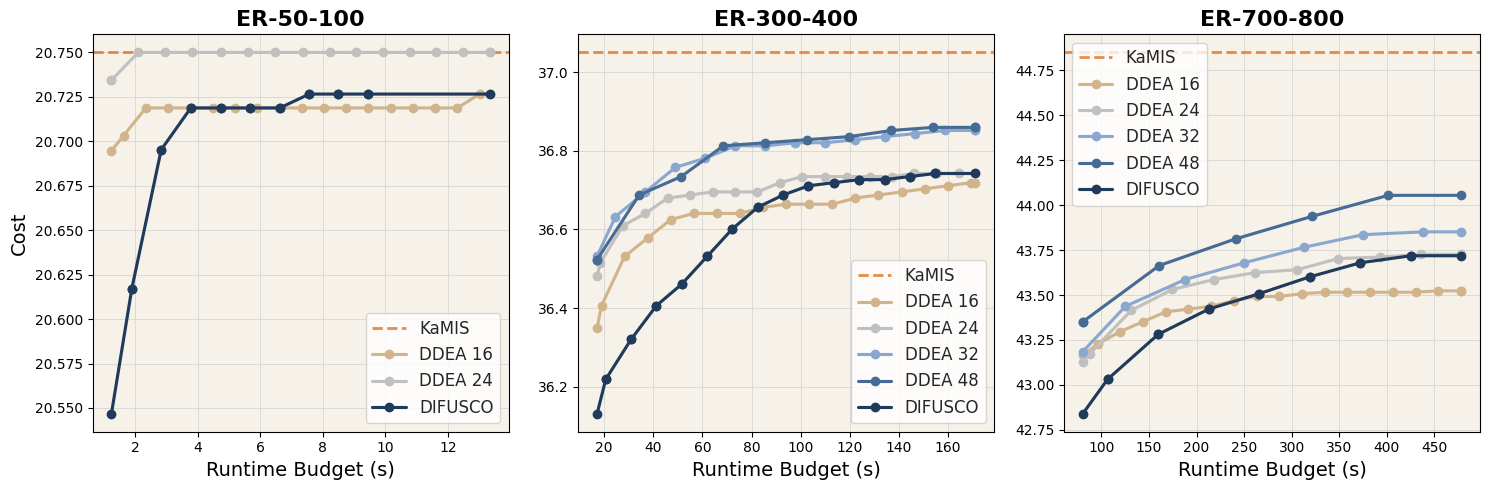}
    \caption{Average cost (IS size, higher is better) against different runtime budgets for DDEA and DIFUSCO.}
\end{figure}

We start by comparing DDEA against DIFUSCO with different population sizes $P\in \{16, 24, 32, 48\}$. Though these are small values compared to standard EAs, they allow DDEA to operate in reasonable runtimes, and they are sufficient to achieve excellent results. We compare against DIFUSCO sampling x32 solutions in parallel, running it until matching the runtime of DDEA. Figure~\ref{fig:cost_vs_budget} shows the comparison of average independent set (IS) size against different runtime budgets for DDEA and DIFUSCO. For DDEA, each plotted datapoint is the average of the cost and the average of the runtime at each generation. For DIFUSCO, each datapoint is the average of the cost and the average runtime at the end of a diffusion sampling call. 
A consistent pattern across datasets shows that DDEA achieves superior solution quality in the early generations. However, with certain suboptimal parameter settings, DIFUSCO's multi-sampling approach can eventually outperform DDEA after extended runtime.
On ER-50-100, two generations are enough for DDEA (P=24) to achieve an optimal solution. In this case, DDEA outperforms DIFUSCO on a time budget of 13 seconds. On ER-300-400, both DDEA (P=16, P=24) are overtaken by DIFUSCO after a number of generations. However, DDEA (P=32, P=48) are able to outperform DIFUSCO on all the considered time budgets. Interestingly, P=32 seems to be the optimal population size for this dataset, as increasing it to P=48 does not improve the results. Finally, on ER-700-800, DDEA (P=32, P=48) are again able to outperform DIFUSCO on all the considered time budgets, but this time, larger population sizes do improve the costs.
On the question of whether DIFUSCO could outperform DDEA with even longer time budgets, we argue that, even if theoretically possible, it would require impractical computational time in the current experimental setup. 

Related to Figure~\ref{fig:cost_vs_budget}, Table~\ref{tab:primal_integral} shows the primal integral for the different methods, which is a runtime-agnostic metric. As mentioned above, the optimal configurations seem to be $P=16, 32, 48$ for ER-50-100, ER-300-400, and ER-700-800 respectively. These configurations significantly outperform DIFUSCO.

\begin{table}[t!]
    \centering
    \renewcommand{\arraystretch}{0.86}
    \caption{Comparison of primal integral (\%) across datasets.}
    \label{tab:primal_integral}
    \setlength{\tabcolsep}{5pt}
    \begin{tabular}{lrrr}
        \toprule
        \multirow{2}{*}{Method} & \multicolumn{3}{c}{Dataset / Primal Integral (\%)} \\
        \cmidrule(lr){2-4}
        & ER-50-100 & ER-300-400 & ER-700-800 \\
        \midrule
        DDEA (P=16) & 0.15 & 1.10 & 3.12 \\
        DDEA (P=24) & \textbf{0.00} & 0.94 & 2.81 \\
        DDEA (P=32) & -- & \textbf{0.70} & 2.62 \\
        DDEA (P=48) & -- & \textbf{0.70} & \textbf{2.27} \\
        DIFUSCO & 0.19 & 1.21 & 3.07 \\
        \bottomrule
    \end{tabular}
\end{table}

After showing the dominance of DDEA over DIFUSCO, independently of the time budget, we now consider a version of DDEA as follows: (P=24, G=15) for ER-50-100, (P=32, G=15) for ER-300-400, and (P=20, G=10) for ER-700-800, and we will denote it as ``DDEA-long'', since it uses relaxed runtime budgets.
We compare against ``DIFUSCO-timed'', which samples 32 solutions in parallel as long as the DDEA-long runs. Additionally, we benchmark DIFUSCO x16, Gurobi solving a standard MISP ILP model (timed to the same runtime as DDEA-long), and KaMIS (timed to 60 seconds).

Table~\ref{tab:benchmark_extended} shows the average cost, the runtime (T), and the number of evaluations (E) and diffusion calls (D) for the different methods.
Note that the number of diffusion calls (D) should be interpreted with caution, as different population sizes (P) result in different batch sizes during sampling, which affects the actual runtime of each diffusion call.
We again observe that DDEA-long outperforms DIFUSCO x16 and DIFUSCO-timed. With respect to the latter one, DDEA provides IS sizes that are 0.25\% and 0.31\% larger on ER-300-400 and ER-700-800 respectively, with slightly shorter runtimes and fewer individuals explored.
DDEA-long also outperforms Gurobi in the ER-300-400 dataset and in the ER-700-800 dataset. DDEA's IS sizes are 3.90\% and 7.50\% larger respectively. Moreover, it matches the costs of both Gurobi and KaMIS on ER-50-100, solving the dataset to optimality.

\begin{table}[t!]
    \centering
    \renewcommand{\arraystretch}{0.86}
    \caption{Comparison of average independent set size (column Cost, higher is better), runtime (column T in seconds), number of function evaluations (column E), and number of diffusion calls (column D) for different MISP solvers.}
    \label{tab:benchmark_extended}
    \setlength{\tabcolsep}{2.5pt}
    \begin{tabular}{l|rrrr|rrrr|rrrr}
        \toprule
        \multirow{2}{*}{Method} & \multicolumn{4}{c|}{ER-50-100} & \multicolumn{4}{c|}{ER-300-400} & \multicolumn{4}{c}{ER-700-800} \\
        \cmidrule{2-13}
        & Cost & T & E & D & Cost & T & E & D & Cost & T & E & D \\
        \midrule
        DIFUSCO x16 & 20.32 & 0.7 & 16 & 1 & 35.53 & 8.7 & 16 & 1 & 42.12 & 43.6 & 16 & 1 \\
        DIFUSCO-timed & 20.73 & 13.9 & 640 & 20 & 36.76 & 207.1 & 960 & 30 & 43.99 & 851.3 & 960 & 30 \\
        DDEA-long & \textbf{20.75} & 13.3 & 540 & 16 & \textbf{36.85} & 183.3 & 720 & 16 & \textbf{44.13} & 803.1 & 720 & 11 \\
        \midrule
        KaMIS (60s) & 20.75 & 60.0 & - & - & 37.05 & 60.0 & - & - & 44.85 & 60.0 & - & - \\
        Gurobi & 20.75 & 0.1 & - & - & 35.47 & 171.2 & - & - & 41.06 & 803.6 & - & - \\
        \bottomrule
    \end{tabular}
\end{table}

\subsection{Out-of-Distribution Generalization}
\label{sec:ood}

Finally, we test DDEA's generalization on the ER-1300-1400 dataset, which is larger than the training instances (with a maximum of 800 nodes). Both DDEA and DIFUSCO use diffusion models trained on ER-700-800.
Table~\ref{tab:ood} compares DIFUSCO (x16 and timed mode) against DDEA (P=16, G=10) on ER-1300-1400, showing average cost, runtime (T), evaluations (E), and diffusion calls (D). DIFUSCO-timed gains only +1.62 with respect to DIFUSCO x16 despite 839 extra seconds runtime. With similar runtime, DDEA achieves notably higher costs. The primal integrals (PIs) confirm this: DDEA's PI (2.27) exceeds DIFUSCO's (1.21) with fewer evaluations and diffusion calls.
This shows DDEA effectively uses evolutionary search to find solutions beyond the training distribution.

\begin{table}[t!]
    \centering
    \renewcommand{\arraystretch}{0.86}
    \caption{Comparison of average independent set size (column Cost, higher is better), runtime (column T in seconds), number of function evaluations (column E), and number of diffusion calls (column D) on the out-of-distribution dataset ER-1300-1400.}
    \label{tab:ood}
    \setlength{\tabcolsep}{3pt}
    \begin{tabular}{lrrrrr}
        \toprule
        Method & Cost & T & E & D \\
        \midrule
        DIFUSCO x16 & 37.07 & 108.1 & 16 & 1 \\
        DIFUSCO-timed & 38.69 & 947.1 & 384 & 12 \\
        DDEA (P=16, G=10) & \textbf{43.17} & 909.5 & 240 & 21 \\
        \midrule
        KaMIS (60s) & 49.91 & 60.0 & -- & -- \\
        \bottomrule
    \end{tabular}
\end{table}

\section{Conclusion}
\label{sec:conclusion}

We introduced DDEA, a framework integrating pre-trained denoising diffusion models into evolutionary algorithms for combinatorial optimization. DDEA leverages diffusion models in two ways: (1) Diffusion-based initialization uses a pre-trained DDM to sample a diverse, high-quality initial population. (2) Diffusion-based recombination employs a novel operator, trained via imitation learning using an ILP-based expert demonstrator, to recombine parent solutions into promising offspring.

Experimental results on the MISP demonstrate DDEA's efficacy. When given the same runtime budget, DDEA outperforms both DIFUSCO and Gurobi. Moreover, DDEA exhibits better out-of-distribution generalization compared to DIFUSCO, showcasing robustness gained from evolutionary search. Ablation studies confirm that both diffusion initialization and recombination components are crucial, with recombination being particularly impactful.

Denoising diffusion-based approaches may scale better to larger problems than autoregressive ML techniques, as the latter are inherently sequential. Despite our promising results, our DDEA implementation faces limitations. The main bottleneck is computational cost; the diffusion inference steps, especially for recombination, dominate the runtime and require GPU resources, hindering scalability. A reason for this is that the current implementation suffers from inefficient in-GPU parallelization. Better GPU utilization could yield significant improvements. Additionally, generating training data for recombination via imitation learning is expensive, requiring EA runs with costly ILP solver calls.

This opens many opportunities for future work. We can optimize diffusion inference parallelization for substantial runtime improvements. Exploring cheaper methods for generating recombination training data, perhaps via a baseline metaheuristic, is crucial for practicality. Applying DDEA to other CO problems appears highly promising, as the framework is generic. Investigating diffusion-based mutation operators and integrating newer, more powerful diffusion backbones (e.g., T2T \cite{li2023t2t}, DISCO \cite{yu2024disco}) could further enhance performance.

In conclusion, DDEA highlights the potential of combining diffusion models' generative strengths with EAs' robust search capabilities. This hybrid ML-metaheuristic approach offers a promising direction for developing effective CO solvers, bridging the gap between pure learning-based methods and specialized classical algorithms.

\section*{Data and Code Availability}
The code and all problem instances used in this work are available at \url{https://github.com/jsalvasoler/difusco_ddea}. 

\bibliographystyle{unsrtnat}
\bibliography{lod_main}

\end{document}